\documentclass{article}

\usepackage{arxiv}

\usepackage[utf8]{inputenc} 
\usepackage[T1]{fontenc}    
\usepackage{hyperref}       
\usepackage{url}            
\usepackage{booktabs}       
\usepackage{amsfonts}       
\usepackage{nicefrac}       
\usepackage{microtype}      
\usepackage{lipsum}
\usepackage{graphics}
\usepackage{graphicx}
\usepackage{mathptmx}
\usepackage{mathtools}
\usepackage{amsmath}
\usepackage{amsfonts,amssymb,amsthm}

\title{Domain-Adversarial Training of Self-Attention Based Networks for Land Cover Classification using Multi-temporal Sentinel-2 Satellite Imagery}

\author{
 Mauro ~Martini\\
  Department of Electronics and Telecommunications\\
  Politecnico di Torino\\
  10124 Turin, Italy\\
  \texttt{mauro.martini@polito.it} \\
   \And
 Vittorio ~Mazzia\\
  Department of Electronics and Telecommunications\\
  Politecnico di Torino\\
  10124 Turin, Italy\\
  \texttt{vittorio.mazzia@polito.it} \\
   \And
  Aleem ~Khaliq\\
  Department of Electrical Engineering\\
  International Islamic University\\
  Islamabad 44000, Pakistan\\
  \texttt{aleem.khaliq@iiu.edu.pk} \\
   \And
 Marcello ~Chiaberge\\
  Department of Electronics and Telecommunications\\
  Politecnico di Torino\\
  10124 Turin, Italy\\
  \texttt{marcello.chiaberge@polito.it} \\
}

\begin{document}
\maketitle

\begin{abstract}
The increasing availability of large-scale remote sensing labeled data has prompted researchers to develop increasingly precise and accurate data-driven models for land cover and crop classification (LC\&CC). Moreover, with the introduction of self-attention and introspection mechanisms, deep learning approaches have shown promising results in processing long temporal sequences in the multi-spectral domain with a contained computational request. Nevertheless, most practical applications cannot rely on labeled data, and in the field, surveys are a time-consuming solution that pose strict limitations to the number of collected samples.  Moreover, atmospheric conditions and specific geographical region characteristics constitute a relevant domain gap that does not allow direct applicability of a trained model on the available dataset to the area of interest. In this paper, we investigate adversarial training of deep neural networks to bridge the domain discrepancy between distinct geographical zones. In particular, we perform a thorough analysis of domain adaptation applied to challenging multi-spectral, multi-temporal data, accurately highlighting the advantages of adapting state-of-the-art self-attention-based models for LC\&CC to different target zones where labeled data are not available. Extensive experimentation demonstrated significant performance and generalization gain in applying domain-adversarial training to source and target regions with marked dissimilarities between the distribution of extracted features.
\end{abstract}


\section{Introduction}
In the past few decades, the launch of many satellite missions with short revisit time and comparatively high-resolution sensors has offered an extensive repository of remote sensing images. Availability of the open-source data by many Earth-observation satellites has made remote sensing very easy and obtainable~\cite{rudd2017application}. Open-source data sets are available free of cost from several satellite missions such as the Sentinel-2 and Landsat~\cite{ novelli2016performance}. These satellites are equipped with multi-spectral sensors with short revisit time, and~good spatial and spectral resolution, allowing researchers to test modern image analysis techniques to extract more detailed information of the target object. It is quite possible to monitor the dynamic processes on Earth~\cite{de2011analysis, pacifici2014importance}. Additionally, it has become easier to estimate and classify biophysical parameters using several data sources~\cite{amoros2013multitemporal, li2015airborne, rembold2013using}. Overall, the~new scenario has led to the opportunity {for} the land cover monitoring, change detection, image mosaicking, and~large-scale processing using multi-temporal and multi-source images~\cite{rudd2017application, khaliq2019refining, gomez2016optical, khaliq2018analyzing}.

The most essential and critical remote sensing application is land cover and crops classification (LC\&CC). It facilitates labeling the cover such as forest, ocean, and~agricultural land. Moreover, mapping can also be done manually using satellite images, but~the process is quite tedious, costly, and~time-consuming. Finally, an~exquisite global cover map is not available as yet, but~there is a land cover map with the name Corine Land Cover (CLC)~\cite{buttner2004corine} which provides land cover information with 100m per pixel resolution. However, the~problem with this map is that it only covers the European area and is updated once in six years. 
There are several ways to perform land classification automatically. In~general, the~ classification involves the creation of a training dataset that consists of annotated samples of the corresponding class labels, training a model using the training dataset, and~evaluating the resulting predictions. The~number and quality of training samples play a pivotal role in defining the performance of the trained model. From a remote sensing prospective, training sample collection requires a ground survey or visual photo-interpretation by an expert~\cite{tuia2016domain}. Ground surveying involves GIS expert knowledge, human resource that is not typically economical, while visual interpretation is not appropriate to be used for some applications, such as finding chlorophyll concentration~\cite{verrelst2012gaussian} and classification of tree species~\cite{ballanti2016tree}. Most of the machine learning (ML) algorithms such as random forest, support vector machines, logistic regression performs well in the context of classification of remote sensing images. However, performance of these ML algorithms are not satisfied when learning features from different sources such as active and passive sensors~\cite{huang2020deep}. It was shown in~\cite{ penatti2015deep, mazzia2020improvement} that Convolutional Neural Networks (CNN) are better than traditional land cover classification {techniques}. In~the land segmentation section of the deep globe challenge~\cite{ demir2018deepglobe}, the~Deep Neural networks (DNN) completely dominate the leaderboards. The~best examples of land cover classification using Deep Neural Networks are ResNet and DenseNet~\cite{tian2018dense, kuo2018deep}.

Since there is a difference in the land covers of different locations, the~model trained in one area cannot be deployed for the other areas. Additionally, the~satellite imagery of different satellites is not the same. That phenomenon is due to the difference in their resolution, capture time, and~other radiometric parameters. Due to these multiple changing variables, the~dataset taken from a satellite covering one region and another satellite dataset covering the same or other regions leads to a domain shift between the datasets. One way to achieve a reliable outcome is possibly to train a model with a huge amount of training samples to generalize its behavior for all classes of all the regions. However, that needs an enormous labeled dataset that is time and labor-intensive.

Another method to deal with the shift between the datasets is termed Domain Adaptation (DA), {in which a model is trained on one dataset (source data) and predictions are made on the other dataset (target domain)}. The~distribution shift between the target and source dataset is mainly due to temporal differences in the acquisition, differences in the acquisition sensors, and~geographical differences {such as variations of objects at the Earth's surface}. The~domain shift affects the performance of a model trained on a source dataset and applied on the target dataset. Domain adaptation methods often rely on learning domain-invariant models that keep comparable performances on the two datasets. Existing domain adaptation techniques may be classified as supervised, unsupervised, and~semisupervised. In~supervised DA methods, it is presumed that labeled data are available {for both source and target domains} \cite{conjeti2016supervised}. In~a semisupervised domain, the~labeled data for the target domain is assumed to be small while an unsupervised method contains labeled data for the source domain only.{ For example in} \cite{7729613}{, a~semisupervised visual domain adaptation was proposed to address classification of very high-resolution remote sensing images. To~deal with the variation in features distribution between the source and target domains, multiple kernel learning domain adaptation method was employed. Another example} \cite{7027189}{, in~which domain adaptation based on semisupervised transfer component analysis was employed to extract features for knowledge transfer from source image to target image for land cover classification of remotely sensed images.}

{Tuia et al.} divides the domain adaptation methodologies into four different categories: domain-invariant feature selection, adapting data distribution, adapting classifiers, and~adaptive classifiers using active learning methods~\cite{tuia2016domain}. {Many studies discuss the unsupervised domain adaptation in the context of classification and segmentation of the remotely sensed satellite and aerial imagery. For~example, In }\cite {liu2020novel}{, an~unsupervised adversarial domain adaptation method was proposed based on boosted domain confusion network (ADA-BDC) which focuses on feature extraction to enhance the transferability of classifier which is trained by source domain images and tested on target domain images. In}~\cite{benjdira2019unsupervised}{, an~unsupervised domain adaptation was used using generative adversarial networks (GANs )for semantic segmentation of aerial images. A~multi-source domain adaptation (MDA) for scene classification was proposed to transfer knowledge from the multiple-source domains to the target domain} \cite{karimpour2020multi}.{ Most of the studies presented in the literature related to DA-based classification have used single date images of source and target domain. However, in~}\cite{bahirat2011novel}, {first approach was proposed in the context of DA for classification of multi-temporal satellite images in which Bayesian classifier-based DA was employed with only two images of Landsat-5 satellite.}

\begin{figure}[ht]
\centering
\includegraphics[scale=0.5]{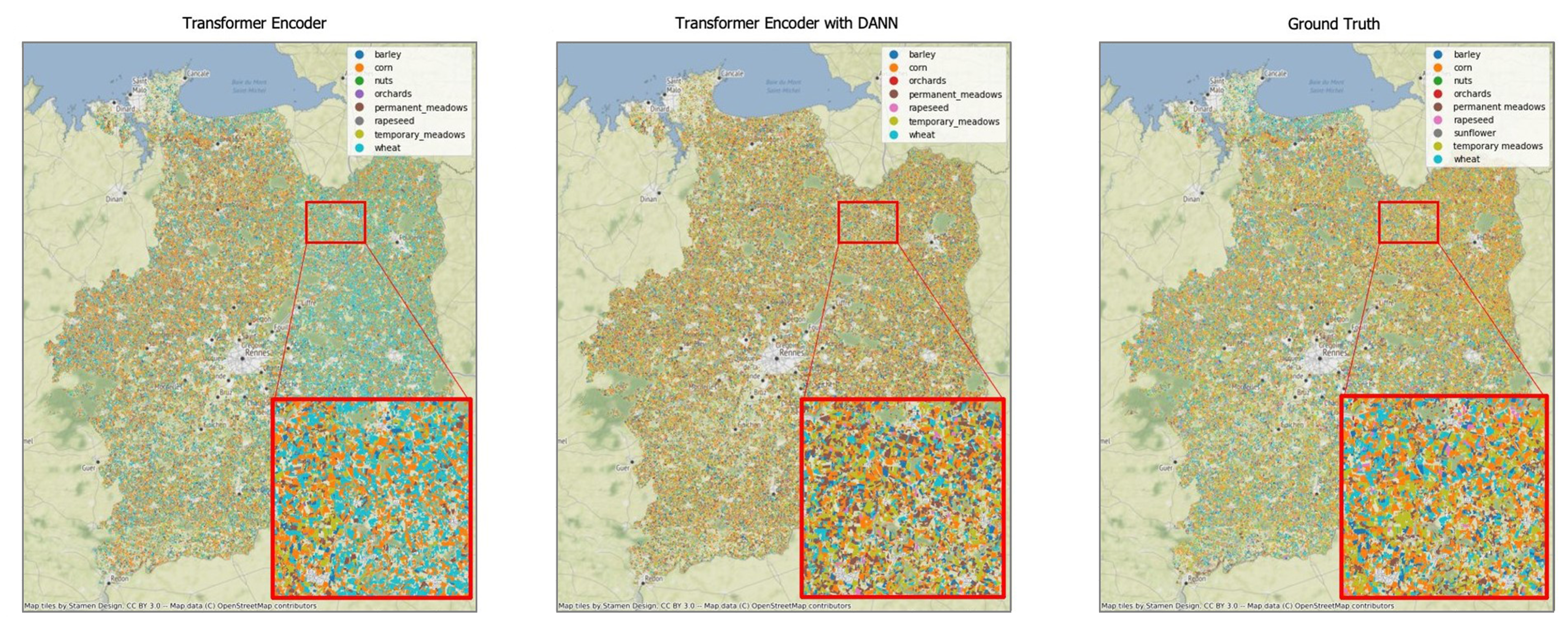}
\caption{Visual representation of land crops classification on zone 3 (Ille-et-Vilain) of the BreizhCrops dataset. For~each sub-image we show the complete region and a sub-area to facilitate the visualization of the advantage obtained by the proposed methodology. In~particular, on~the left the crops predictions without our domain adaptation mechanism are shown, while in the center the same predictions performed adopting DANN are proposed. On~the right, ground truth labeled crops can be visualized. The~improvement in the classification with DANN is evident, especially in the reduction of misclassification of wheat and~meadows.}
\label{fig:map_zone3}
\end{figure}

This work investigates adversarial training of deep neural networks to bridge the domain discrepancy between distinct geographical zones. In~particular, we perform a thorough analysis of domain adaptation applied to challenging multi-spectral, multi-temporal data, highlighting the advantages of adapting state-of-the-art self-attention-based models for LC(\&)CC to different target zones where labeled data are not available. We choose to experiment our methodology on the BreizhCrops dataset, a~large-scale time series benchmark dataset introduced in 2020 by Rußwurm~et~al., \cite{russwurm2019breizhcrops}, for~supervised classification of field crops from satellite data. Figure~\ref{fig:map_zone3} shows the visual representation of the crop prediction performed on a sub-region of Brittany, highlighting the benefit provided by the proposed~methodology.

This article is organized as follows. Section~\ref{related_work} covers the related work on domain adaptation and its developments in techniques for LC\(\&\)CC. Section~\ref{study_area} describes the dataset. A~detailed description of the proposed method is presented in Section~\ref{methodology}. The~experimental setup, the~results and related discussion are reported in Section~\ref{experiments}. Finally, Section~\ref{conclusion} draws some conclusions and future~directions.

\section{Related~Work}
\label{related_work}
\subsection{Land Cover and Crop~Classification}
LC\&CC has been the subject of many studies in the past. A~widely used classification method makes use of time series of vegetation indices {(VI)} derived from remotely sensed imagery to extract temporal features and phenological metrics. There are also some thresholds and simple statistical techniques that help calculate the time of peak VI, Maximum VI, and~other vegetation related metrics~\cite{walker2014dryland,arvor2011classification}. {Moranduzzo et al. and Hao \mbox{et al.}} \cite{ moranduzzo2013automatic, hao2015feature} Illustrate the older image classification methods using handcrafted features for image representation and training classifiers such as support vector machine and random forest. {Machine learning} methods self-learn how to extract the features from the data with massive datasets available and improved computing devices. Random Forest (RF)-based classifiers is another common approach for remote sensing applications~\cite{hao2015feature}, though~it should be noted that multiple features need to be derived and fed to the RF classifier for more effective~output. 

One of the newest and most powerful concepts integrated into mapping is a branch of machine learning known as Deep Learning (DL). {DL is a type of machine learning based on artificial neural networks in which multiple layers of processing are used to extract progressively higher-level features from data}. DL can be used to solve a wide range of problems such as signal processing, computer vision, image processing, and~natural language processing~\cite{lecun2015deep}. DL has shown significant contribution in remote sensing image classification due to its ability to represent features and its competence of mechanization for end-to-end learning. Autoencoders are type of artificial neural networks and are often used to represent features of data~\cite{ hubel1962receptive, szeliski2010computer}. In~the remote sensing field, object detection and image segmentation have been performed extensively using two-dimensional CNNs~\cite{krizhevsky2012imagenet, zeiler2014visualizing} to perform spatial feature extraction from high-resolution images. 2D CNN proved better than 1D CNN in crop classification~\cite{ kussul2017deep}.
{In remote sensing, two-dimensional CNN can be used effectively for image classification where the correlation between the morphological details and the target classes exists. For~example in }\cite{zhang2016deep},{ a 2D-CNN is used to obtain the spatial features of the hyperspectral imagery (HSI), analyzing the continuity of land covers in the spatial domain. Often relation among spectral bands of HSI is not linear, in~that case, 2D-CNNs are normally used together with 1D-CNNs to incorporate the spectral and spatial domain of features }\cite {zhang2017spectral}.

{Indeed, the~classification task becomes quite challenging when dealing with high-dimensional hyperspectral data with few labeled samples. Recently, generative adversarial networks (GANs) have been exploited for sample generation, though~it is not easy to acquire high-quality samples with authenticity. In this context, the~generative adversarial networks (GANs) aim to generate more labeled samples by mimicking labeled data and provide high-quality realistic data to increase the number of training samples} \cite{he2017generative}. {Generally, GANs are comprised of two adversarial modules: a generator that obtains the original data distribution and a discriminator that differentiates between the generated labeled data and the original ones }\cite{goodfellow2014generative}.{ For this purpose, an~unsupervised 1D GAN was aimed to capture the spectral distribution while increasing the training samples for HSI classification }\cite{zhan2017semisupervised}. {It was trained on unlabeled samples, which were then transformed as a classifier in a semisupervised setting. Hence, it is difficult to learn class features during the training process. Modified versions of GANs have considered the label information, such as conditional GAN (CGAN) }\cite {isola2017image}, InfoGAN~\cite{chen2016infogan},{ deep convolutional GAN (DCGAN)} \cite{radford2015unsupervised},{ and categorical GAN (CatGAN) }\cite{springenberg2015unsupervised}.

{The aforementioned versions of GANs are susceptible to noises and disregard the relationships between spectral bands. Additionally, the~generated samples are usually very different in the spectral domain from the original ones which fail in increasing classification results. This problem has been addressed in} \cite{zhao2020sample},{ authors developed a self-attention generative adversarial adaptation network (SaGAAN) to produce high-quality labeled samples in the spectral domain for hyperspectral image classification.}

\subsection{Domain~Adaptation}
The method of domain adaptation aims to reduce the domain shift between source and target datasets. Domain adaptation has three possible approaches according to~\cite{ wang2018deep, ma2019super, bengana2020improving}. The~primary approach consists of reducing the difference in the feature space among the target and source data. For~this purpose, maximum mean discrepancy (MMD) is often used as a cost function to minimize the distance or to check a consistent feature extraction in both source and target domains~\cite{ bengana2020improving}. {Other investigations focus on feature extraction; however,  Nielsen et al.} \cite {nielsen2007regularized}{ performed change detection by aligning both domains using canonical correlation analysis (CCA). The~work is extended with a semisupervised approach, where change detection is performed on multi-scale data obtained from different sensors }\cite{volpi2015spectral}. In~\cite{tuia2014semisupervised},{ the domain alignment is achieved through an eigenproblem aiming at preserving the mismatch of labels and the geometric structure.} 
The second approach uses Generative Adversarial Networks (GANs) \cite{ goodfellow2014generative} for an adversarial domain adaptation. The~purpose of the GANs is to make both the source and target datasets spectral characteristics similar. {Tzeng et al.} \cite{tzeng2017adversarial} shows an example where the target dataset is translated to the source dataset using GANs. The~translation contains a discriminator that {recognizes} the two datasets. {Most of the studies employ a feature extraction network to generate feature sets for source and target domain} \cite{hosseini2018augmented,volpi2018adversarial,taigman2016unsupervised}. {The feature extraction network acts as a generator to reduce the classification loss for the source domain and concurrently maximize the loss of the discriminator. Based on these approaches, Adversarial Discriminative Domain Adaptation (ADDA)  was employed to learn feature extraction networks for the source as well as for target domains }\cite{tzeng2017adversarial}. In~\cite{volpi2018adversarial},{ an adversarial feature augmentation method was proposed to achieve DA in which the encoder is trained for the source and target domains. Inspired by the concept used in ADDA, Mesay et al.} \cite{bejiga2018gan} {implemented GAN-based DA for object classification in the remote sensing data.} The last approach of domain adaptation creates a shared representation of both domains. In~this method, one domain can be translated to another, and~both domains can be translated into a common space. The~method also provides a transfer function that facilitates the translation of one domain to another and translating back to the original state. CycleGAN provides the third approach and involves two discriminators that are used to translate one domain to another and converse~\cite{zhu2017unpaired}.

 The general methods of domain adaptations are not well interpreted for semantic segmentation~\cite{zhang2017curriculum}. Thus, adversarial and reconstruction procedures are chosen. Adversarial and constraint-based adaptations are performed at pixel level using architectures that exploit adversarial domain adaptation using GANs to {transform} source-like images~\cite{hoffman2016fcns}. Then, the~images are segmented using a network that has been trained on the source dataset. In~\cite{chang2019all}, Domain-Invariant Structure Extraction (DISE) structure was adopted to transform images into the domain-invariant structure and domain-specific texture representations. The~bidirectional method {prevents} the translational model to reach a point where the discriminator fails to identify the image from the same distribution setup and fails to align correctly~\cite{li2019bidirectional}.

\section{Study Area and~Data}
\label{study_area}
\begin{figure}[ht]
\centering
\includegraphics[scale=0.38]{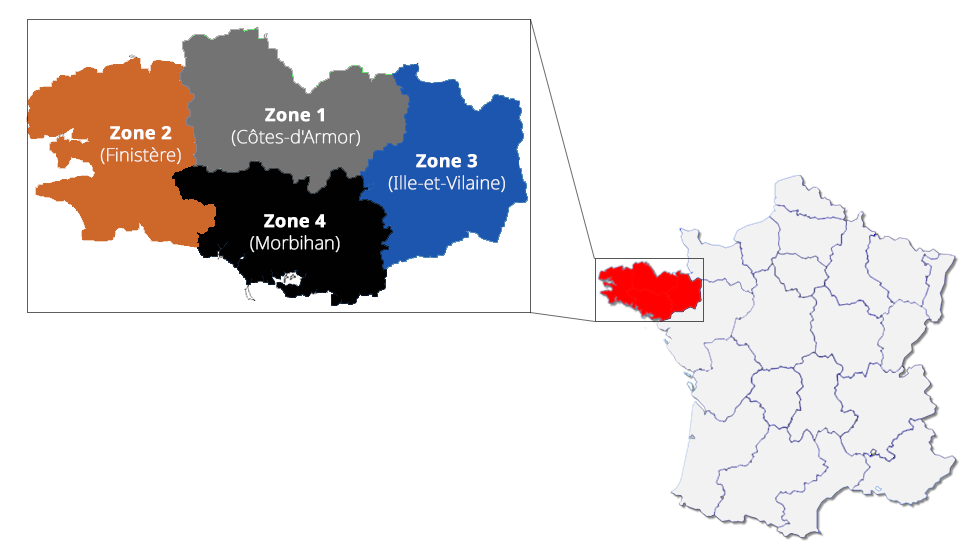}
\caption{Magnified view of the four NUTS-3 regions of Brittany, located in the northwest of France and covering 27,200 km². The~strict division of the supervised BreizhCrops dataset in the four regions allows the performance of a formal and controlled analysis on domain adaptation for LC\&CC with multi-spectral and multi-temporal~data.} 
\label{fig:map}
\end{figure}

To promote reproducibility of our experimentation, we rely on BreizhCrops, a~large-scale time series benchmark dataset introduced in 2020 by Rußwurm~et~al., \cite{russwurm2019breizhcrops}, for~supervised classification of field crops from satellite data. The~dataset comprises multivariate time series examples in the Region of Brittany, France, of~the season 2017, from~January 1 to December 31. In~particular, the~authors of the dataset exploited all available Sentinel 2 images from Google Earth Engine,~\cite{gorelick2017google}, and~farmer surveys collected by France National Institute of Forest and Geography Information (IGN) to collect more than 600 k samples divided into 9 classes with 45 temporal steps and 13 spectral bands. Most importantly, as~shown in Figure~\ref{fig:map}, acquired data are {equally} split into distinct regional areas. Indeed, as~regulated by the Nomenclature des unites territoriales statistiques (NUTS), the~overall dataset is divided into the four NUTS-3 regions Côes-d’Armor, Finistère, Ille-et-Vilaine, and~Morbihan. That, in~conjunction with the challenging nature of the dataset, makes BreizhCrops an ideal benchmark to test domain adaptation for multi-spectral and multi-temporal data for LC\&CC.

\begin{figure}[h]
\centering
\includegraphics[scale=0.18]{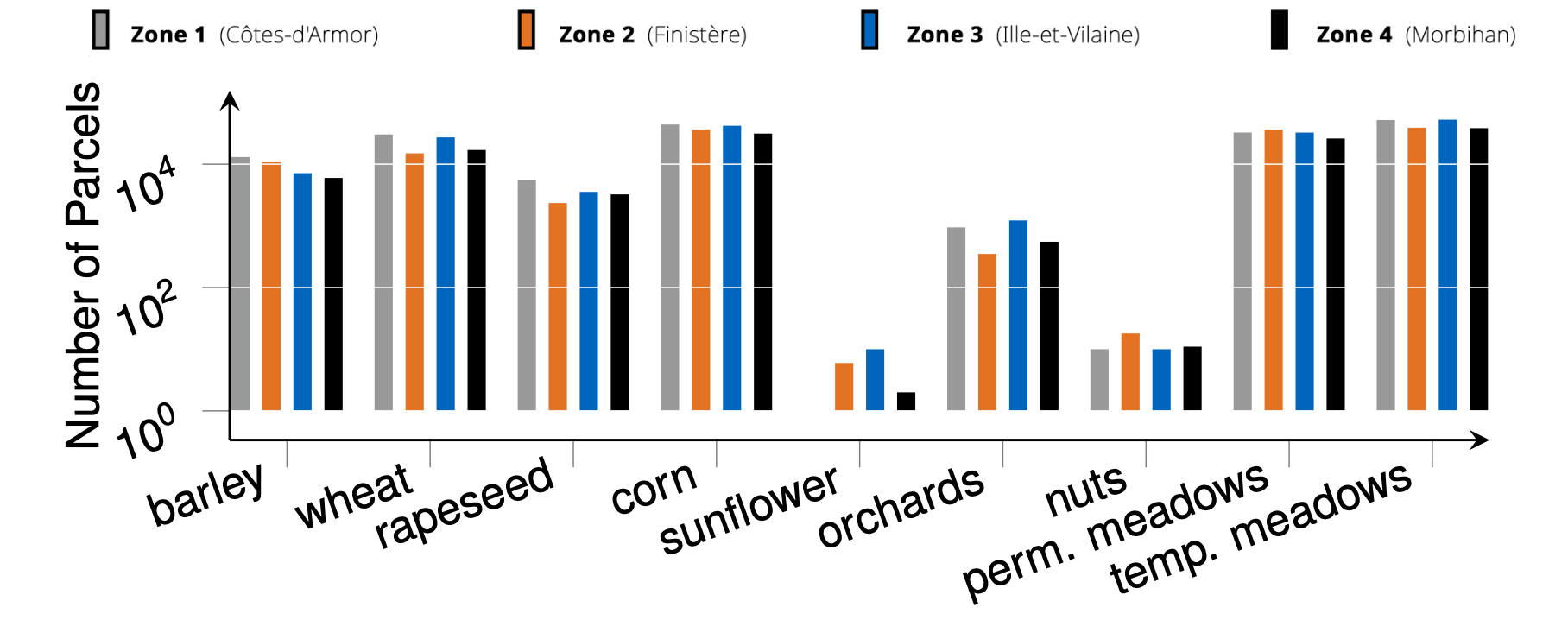}
\caption{Class frequencies divided in the four NUTS-3 regions of Brittany. The~respective number of parcels highlights the strong class imbalance, reflecting the substantial imbalance in real-world crop-type-mapping datasets. {However, samples per class in the four regions are equally divided.}}
\label{fig:n_parcells}
\end{figure}

As summarized in Figure~\ref{fig:n_parcells}, even if the authors of the dataset avoided broad categories, due to the nature of agricultural production, which focuses on a few dominant crop types, a~class imbalance can be observed in the collected parcels. That constitutes a challenge for every classifier type, but~it reflects the strong imbalance in real-world crop-type-mapping datasets. {On the other hand, sample classes in the different regions are balanced, making BreizhCrops a perfect bench for testing domain adaptation strategies.}
Finally, to~disentangle the performed domain adaptation analysis from the influence of the random variation of the atmospheric conditions, we exclusively make use of L2A bottom-of-atmosphere {imagery} where data acquired over time and space share the same reflectance scale. Adjacent and slope effects are corrected by the MAJA processing chain~\cite{hagolle2015multi} that employs 60-meter spectral bands to apply atmospheric rectification and detect clouds. Therefore, only ten spectral features are available for each parcel. Table~\ref{tab:classe_summary} is presented as a summary of the number of samples collected for the domain adaptation experimentation divided into classes and regions. In conclusion, multi-spectral, multi-temporal pixels are individually extracted for each parcel and are constituted by 10 spectral bands and 45 temporal steps each. The~class imbalanced highlighted by the number of parcels of Figure \ref{fig:n_parcells}{ is reflected in the number of samples of Table} \ref{tab:classe_summary} {used for all experimentation.}

\begin{table}[h]
\caption{Summary of the number of samples per class divide in the four NUTS-3 regions of Brittany. Instances are derived by L2A bottom-of-atmosphere parcels to disentangle our analysis with variation of the atmospheric~conditions.}
\label{tab:classe_summary}
\setlength{\tabcolsep}{2.7mm}{  
\begin{tabular}{llllllllll}
\toprule
       & \textbf{Barley} & \textbf{Wheat} &\textbf{ Rapeseed} & \textbf{Corn}  & \textbf{Sunflower} & \textbf{Orchards} & \textbf{Nuts} & \begin{tabular}[c]{@{}l@{}}\textbf{Permanent} \\ \textbf{Meadows}\end{tabular} & \begin{tabular}[c]{@{}l@{}}\textbf{Temporary}\\  \textbf{Meadows}\end{tabular} \\ \midrule
Zone 1 & 13,051  & 30,380 & 5596     & 44,003 & 1         & 937      & 10   & 32,641                                                           & 52,013                                                        \\ 
Zone 2 & 10,736  & 15,026 & 2349     & 36,620 & 6         & 348      & 18   & 36,536                                                        & 39,143                                                        \\ 
Zone 3 & 7154   & 27,202 & 3557     & 42,011 & 10        & 1217     & 10   & 32,524                                                        & 52,682                                                        \\ 
Zone 4 & 5981   & 17,009 & 3244     & 31,361 & 2         & 552      & 11   & 26,134                                                        & 38,141                                                        \\ \bottomrule
\end{tabular}}
\end{table}

\section{Methodology}
\label{methodology}
In this work, unsupervised domain adaptation is considered in the field of land cover classification from satellite images. {The study aims to tackle the problem of low generalization capability of classifiers only trained on a peculiar geographical region dataset. Moreover, the~lack of rich available datasets of labeled satellite images increases the interest towards this challenge.} In particular, the~proposed methodology is intended to investigate the application of representation learning (RL) techniques for domain adaptation when dealing with multi-temporal data. For~this purpose, a~Transformer Encoder-based classifier is adapted to a Domain-Adversarial Neural Networks (DANN) architecture and trained~accordingly.

In this section, a~thorough description of the methodology is provided. First, we frame domain adaptation with the DANN method. Then, we briefly explain the Transformer Encoder structure with self-attention adopted for the multi-temporal crops classification. Finally, we describe the resulting architecture of the attention-based DANN, which is used to train a classifier with improved domain~generalization.

\subsection{Domain-Adversarial Neural~Networks}
Classifiers obtained with Deep Neural Networks often suffer from a lack of generalization related to possible variations in the appearance of the same objects. This problem is usually identified as a domain gap. In~the land cover classification task, this situation is very recurrent and can be associated with the spectral shift affecting the data collected in different regions at different times. The~shift is often related to photogrammetric distortion or visual differences in the appearance of lands. Furthermore, when dealing with satellite images, a~dataset usually needs to be created by labeling images for a specific region to train a classification model. Despite this time-expensive procedure, standard training does not guarantee satisfying performance on images of different~regions.

Domain-Adversarial Neural Networks (DANN) is a representation learning technique that allows a classifier to generalize better from a \textit{source domain} to a \textit{target domain}. This specific domain adaptation method consists of adding a branch to the original feed-forward architecture of the classifier and carry out an adversarial training.
From a generic perspective, it is possible to identify three main components of the DANN: a \textit{feature extractor} with parameters \(\theta_{f}\), a~\textit{label predictor} with parameters \(\theta_{y}\), and~a \textit{domain classifier} with parameters \(\theta_{d}\). The~feature extractor is the first block of the DANN model. It is responsible for learning the function \(G_{f}:X \to \mathbb{R}^{d}\), which maps the input samples \(X\) to a d-dimensional vector containing the extracted features. The~label predictor function, \(G_{y}(G_{f}(X))\), compute the label associated with the predicted class of the sample. The~domain discriminator function \(G_{d}(G_{f}(X))\) distinguishes between source and target domains given the extracted features. The~combination of feature extractor and label predictor gives us the complete classifier model. The~domain classifier is composed of a secondary branch, similar to the label predictor, which receives the extracted feature vector by the first block of the~network.

Given these three main elements, the~expression of the total loss used to train DANN is obtained by the following expression, according to the authors~\cite{ganin2016domain}:
\begin{equation}
\label{eqn:loss}
\mathcal{L}\left(\theta_{f}, \theta_{y}, \theta_{d}\right)=\frac{1}{n} \sum_{i=1}^{n} \mathcal{L}_{y}^{i}\left(\theta_{f}, \theta_{y}\right)-\lambda\left(\frac{1}{n} \sum_{i=1}^{n} \mathcal{L}_{d}^{i}\left(\theta_{f}, \theta_{d}\right)+\frac{1}{n^{\prime}} \sum_{i=n+1}^{N} \mathcal{L}_{d}^{i}\left(\theta_{f}, \theta_{d}\right)\right)
\end{equation}

The first term \(\mathcal{L}_{y}\) is the label predictor loss, while the second one involves {the domain discriminator loss} \(\mathcal{L}_{d}\). The~hyper-parameter \(\lambda\) can be tuned to weigh the contribution of the two learning terms. A~more detailed analysis of the choice of \(\lambda\) is proposed in the experiments section. \(n\) and \(n'\) are respectively the numbers of samples from the source and the target domains. {Totally, we have} \(N=n+n^{\prime}\) {samples used in the training.} The expression of the total loss function also describes the principal goals of DANN: first, we want to obtain a label predictor with low classification risk. Second, we are adding a regularization term for the domain adaptation. To~this extent, we aim to find a set of parameters of the feature extractor \(\theta_{f}\) that can map a generic input sample from either source or target domain to a new latent space of features, where the domain gap is reduced. On~the other hand, the~classification performance has not to be affected. For~this reason, the~extracted features should be discriminative as well as domain-invariant.
According to this goal, the~optimal choice of parameters \(\theta_{f}\) and \(\theta_{y}\) is represented by the one which minimizes the total loss function, keeping \(\hat{\theta_{d}}\) unchanged. By contrast, the~domain discriminator parameters \(\theta_{d}\) are updated to maximize the loss while not changing the other~ones.
\begin{equation}
\label{eqn:optimal1}
\left(\hat{\theta}_{f}, \hat{\theta}_{y}\right)=\underset{\theta_{f}, \theta_{y}}{\operatorname{argmin}} \mathcal{L}\left(\theta_{f}, \theta_{y}, \hat{\theta}_{d}\right)
\end{equation}
\begin{equation}
\label{eqn:optimal2}
\hat{\theta}_{d}=\underset{\theta_{d}}{\operatorname{argmax}} \mathcal{L}\left(\hat{\theta}_{f}, \hat{\theta}_{y}, \theta_{d}\right) .
\end{equation}

In the original paper of DANN, the~parameters of each piece of the neural network model are updated with a {classical Stochastic Gradient Descent (SGD) optimizer. Here instead we use Adam (Adaptive momentum estimation), another popular optimization algorithm introduced by} \cite{kingma2014adam}. Parameters \(\theta_{f}\),\(\theta_{y}\) and \(\theta_{d}\) are updated according to its rules.
\begin{equation}
\label{eqn:update1}
\theta_{f} \quad \longleftarrow \quad \theta_{f}-\eta \left(\dfrac{\hat{m}_{f,y}}{\sqrt{\hat{v}_{f,y}} + \epsilon}  -\lambda \dfrac{\hat{m}_{f,d}}{\sqrt{\hat{v}_{f,d}} + \epsilon} \right)
\end{equation}
\begin{equation}
\label{eqn:update2}
\theta_{y} \quad \longleftarrow \quad \theta_{y}-\dfrac{\eta}{\sqrt{\hat{v}_y} + \epsilon} \hat{m}_y
\end{equation}
\begin{equation}
\label{eqn:update3}
\theta_{d} \quad \longleftarrow \quad \theta_{d}-\dfrac{\eta}{\sqrt{\hat{v}_d} + \epsilon} \hat{m}_d
\end{equation}

{As can be studied more in detail in the Adam original paper, the~first (mean) and the second (uncentered variance) moments of Adam} \(\hat{m}\) and \({\hat{v}}\) {are estimated as exponentially moving averages computed with the gradients obtained from each mini-batch. For~the specific case of DANN, gradients used to estimate the Adam moments change for each element} \(G_{f}\), \(G_{y}\), \(G_{d}\) {of DANN structure}. For~example, the~feature extractor gradients
\((\partial \mathcal{L}_{y}^{i}/\partial \theta_{f})\) and \((\partial \mathcal{L}_{d}^{i}/\partial \theta_{f})\) are used to compute \(\hat{m}_{f,y}\) and \(\hat{m}_{f,d}\). {Diversely,} gradients obtained from label predictor \((\partial \mathcal{L}_{y}^{i}/\partial \theta_{y})\) and domain discriminator \((\partial \mathcal{L}_{d}^{i}/\partial \theta_{d})\) are only used to update their respective momentum \(\hat{m}_{y}\) and \(\hat{m}_{d}\).

The feature extractor and the domain discriminator play adversarial roles during the training process. A~satisfying feature extractor can fool the domain discriminator by forwarding a vector of domain-invariant features. The~role of the domain discriminator is to improve and evaluate this ability. A~key intuition in the DANN method is to carry out the adversarial training with a standard backpropagation of the gradients, thanks to a custom Gradient Reversal Layer between the feature extractor and the domain discriminator. This particular layer does not add other parameters to the model but changes the sign of the upstream gradients. The~GRL operation can be formulated with \(\mathcal{R}(\textbf{x})\) in the following mathematical expressions for the forward and backpropagation step:
\begin{equation}
\label{eqn:GRL1}
    \mathcal{R}(\textbf{x}) = \textbf{x}
\end{equation}
\begin{equation}
\label{eqn:GRL2}
\frac{d \mathcal{R}}{d \mathbf{x}}=-\mathbf{I}
\end{equation}
where \(\mathbf{I}\) is the identity matrix. Hence, by~performing optimization steps on the resulting DANN architecture, we can update parameters to reach saddle points of the total loss function reported in~\eqref{eqn:loss}.

\subsection{Classification of Multi-Spectral Time Series Data with~Self-Attention}
Self-Attention, popularized by the Transformer model in 2017,~\cite{vaswani2017attention}, has provided a considerable boost in machine translation performance while being more parallelizable and requiring significantly less time to train. Nevertheless, the~introspection capability behind the success of Transformers is not limited only to natural language processing, but~can be adapted to any time series analysis to filter data and focus on more relevant repressions~aspects.  

A {single sample pixel} $i$-th of multi-spectral, multi-temporal acquisition can be represented as a matrix $\textbf{\textit{X}}^{(i)} \in \mathbb{R}^{t\times b}$ where $t$ is the temporal dimension and $b$ is given by the number of spectral bands. Therefore, it is a 1D sequence of tokens, $(\textbf{\textit{x}}_{0},...,\textbf{\textit{x}}_{t})$, with~$\textbf{\textit{x}}_{t} \in \mathbb{R}^{b}$, that can be easily linearly projected to feed a standard Transformer encoder. The~encoder can map a temporal input sequence $\textbf{\textit{X}}_{t \times b}$ in a continuous representation $\textbf{\textit{X}}^{L}_{t \times d_{model}}$, where $L$ is the output layer of the Transformer model and $d_{model}$ is the constant latent dimension of the projection~space.

Self-attention, through local multi-head dot-product self-attention blocks, can easily manipulate the temporal sequence finding correlations between different time-steps and completely avoiding the use of recurrent layers.
{The dot-product self-attention operation is composed on a trainable associative memory with key and value vector pairs of dimensions} $d$. For~a sequence of $t$ {query vectors, arranged in a matrix} $Q \in \mathbb{R}^{t\times d}$, {the self-attention operation is described by the following operation:}
\begin{equation}
\label{eqn:atten}
\text{Attention}(Q,K,V)=\text{Softmax}(QK^T/\sqrt{d})V
\end{equation}
{where the Softmax function is applied over each row of the input matrix and} $K \in \mathbb{R}^{t\times d}$ and $V \in \mathbb{R}^{t\times d}$ {are the key and value vector matrices, respectively. Query, key and values matrices are themselves computed from a sequence of} $t$ {input vectors with dimension} $d_{model}$ {using linear transformations}: $Q=XW_{Q}, K=XW_{K}, V=XW_{V}$ {where} $X \in \mathbb{R}^{t\times d_{model}}$. {Finally, multi-head dot-product self-attention is defined by considering applying} $h$ {self-attention functions to the input} $X$. {Each head provides a sequence of size} $t \times d$. {These} $h$ {sequences are rearranged into a} $t \times dh$ {sequence that is linearly projected into} $t \times d_{model}$.

Subsequently, {after the transformer encoder}, the~output representation, $\textbf{\textit{X}}^{L}_{t \times d_{model}}$,can be exploited to perform a classification of the input sequence. Indeed, that can be achieved by further processing the output encoder matrix and feeding a classification head trained to map the hidden representation to one of the $k$ classes. 

Several approaches have been proposed in the literature to obtain this result; in~\mbox{\cite{devlin2018bert,dosovitskiy2020image}} they pre-append to the input sequence a learnable embedding, whose state at the output of the Transformer encoder serves as a hidden representation of the membership class. Indeed, only that output token is fed to the classification head to obtain the final prediction. On~the other hand, the~output sequence can be averaged or processed with a max operation on the temporal dimension~\cite{russwurm2020self}. Nevertheless, despite the type of processing applied to $\textbf{\textit{X}}^{L}$, the~encoder will adapt to elaborate the sequence properly and embed the needed information for the classification task. In~conclusion, a~Transformer encoder can be repurposed to process a multi-spectral input sequence and find valuable correlations between the different time-steps to perform LC\&CC with a high level of~accuracy. 

\subsection{DANN for Land Cover and Crop~Classification}

\begin{figure}[ht]
\centering
\includegraphics[scale=0.35]{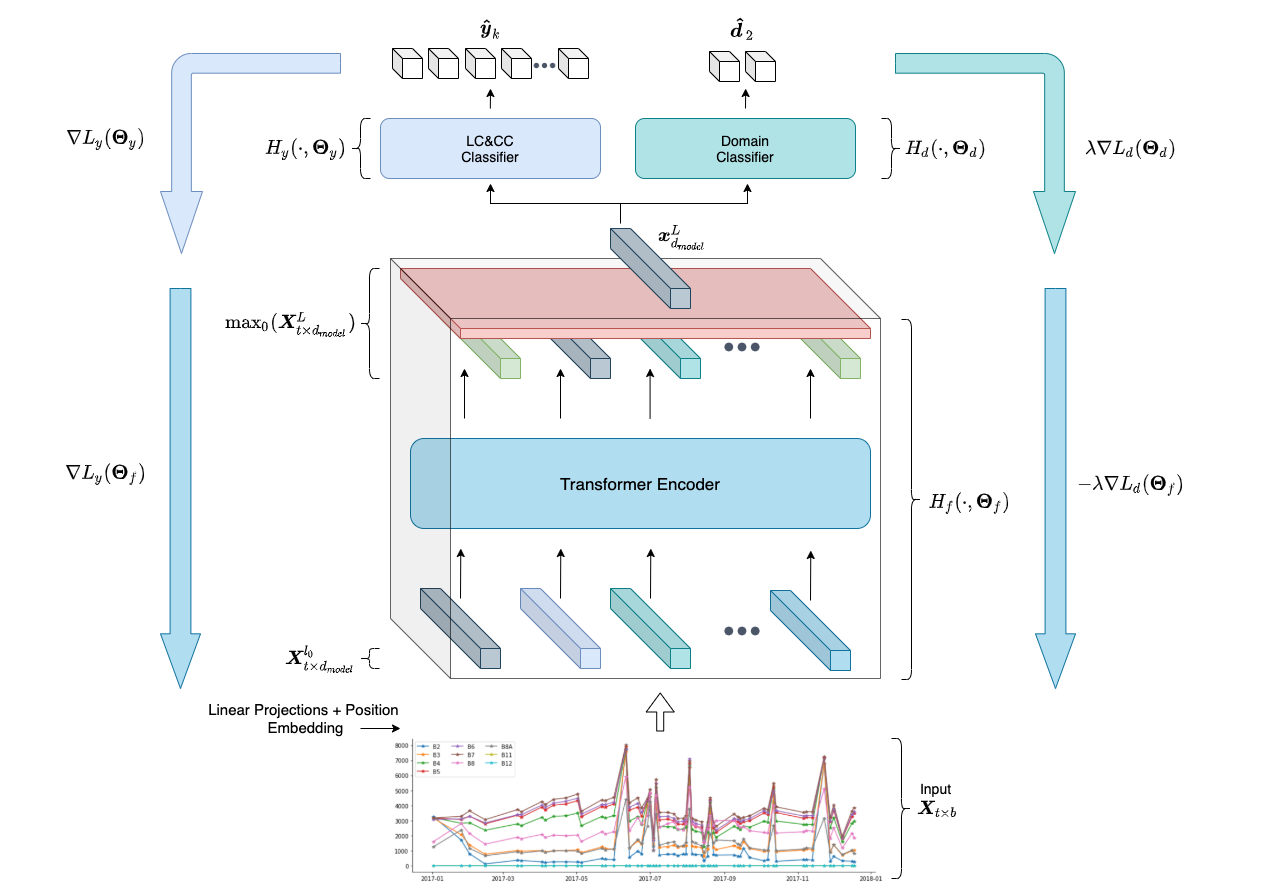}
\caption{Overview of the overall framework to train a Transformer encoder with domain-adversarial training. The~multi-spectral temporal sequence $\textbf{\textit{X}}_{t \times b}$ is first linearly projected and fused with a position encoding. Subsequently, the~self-attention-based model manipulates the input series and, through a max operation applied to the last layer of the encoder, is possible to extract a token $\textbf{\textit{x}}^{L}_{d_{model}}$ from the output sequence. Finally, gradients derived by LC\&CC and Domain classifiers train the network while keeping close the distribution of source and target~domains.}
\label{fig:overall_architecture}
\end{figure}

We employ DANN in conjunction with self-attention-based models to bridge the domain gap between different geographical regions. The~overall architecture of the adopted methodology is shown in Figure~\ref{fig:overall_architecture}. First, an~input sequence $\textbf{\textit{X}}_{t \times b}$ is linearly projected to the constant latent dimension of the Transformer model $d_{model}$. Moreover, a~Transformer encoder does not contain recurrence or convolution to make use of the order of the sequence. Therefore, some positional encoding is injected about the relative or absolute position of the tokens in the sequence. The~positional encodings have dimension $d_{model}$
as the projected sequence, so that the two can be summed. Guided by experimentation, as~in~\cite{dosovitskiy2020image}, we adopt a learnable positional encoding instead of the sine and cosine functions with different frequencies of~\cite{vaswani2017attention}. The~resulting pre-processed input sequence $\textbf{\textit{X}}_{t\times d_{model}}^{l_{0}}$ feeds the Transformer encoder, parameterized by $\Theta_{f}$, that provides as output a continuous representation $\textbf{\textit{X}}^{L}_{t \times d_{model}}$. Subsequently, we make use of the max function, over~the temporal axis, to~extract a token, $\textbf{\textit{x}}^{L}_{d_{model}}$, from~the output~sequence.

The extracted representation constitutes the input for either the LC\&CC and domain multi-layer perceptron classifiers. The~first network provides a probability distribution over the $k$ different classes, $\hat{\textbf{\textit{y}}}_{k}$. On~the other hand, the~domain classifier outputs the probability, $\hat{\textbf{\textit{d}}}_{2}$, that the extracted representation $\textbf{\textit{x}}^{L}_{d_{model}}$ belongs to the target or source domain. Using the cross-entropy loss function for both classifiers, it is possible to compute the respective gradients and update the weights, $\Theta_{f}$ of the feature extractor. Indeed, inverting the sign of the gradients, $\nabla L_{d}(\Theta_{d})$, derived from the domain classifier, and~multiplying them for a scale factor $\lambda$, we can increasingly reduce the distance between the latent space of the two domains while training the encoder on the classification task. Overall, the~proposed training framework provides an effective solution to transfer the acquired knowledge of a model to a diverse region, exploiting only the original nature of the~data.

\section{Experiments and Discussion}
\label{experiments}
We experiment with the proposed methodology on the four regions of the multi-temporal satellite BreizhCrops dataset presented in Section~\ref{study_area}. As~explained in the same section, we indicate this dataset as an optimal choice to train and test new domain adaptation methods exploiting labeled multi-temporal data. The~first main objective of the conducted experimentation is to investigate how the classification performance of a state-of-the-art model for LC\&CC model is affected by a lack of generalization towards different geographical regions. Then, we clearly highlight how adversarial training can mitigate the domain gap and significantly boost performance for source and target regions with marked distribution distance. It is important to remark that the method relies on the availability of samples of both source and target domains, whereas only source labels are required, not allowing direct applicability of transfer learning techniques. Finally, in~the last part of the section, obtained results are discussed and inspected through dimensionality reduction techniques, validating the proposed method for practical~use.

\begin{figure}[h]
\centering
\includegraphics[scale=0.5]{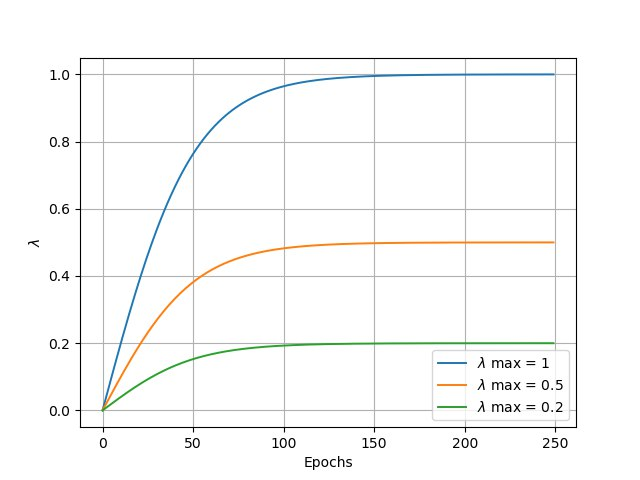}
\caption{\(\lambda\) scheduling: the value of the domain adaptation parameter \(\lambda\) is changed during training according to an exponentially growing trend. This allows the feature extractor to learn basic features during the initial epochs. Different final \(\lambda_{max}\) values are tested to study the right level of adaptation required in the different cases: 1, 0.5 and 0.2. \(\lambda_{max} = 0.2\) is the best choice for an overall adaptation improvement of the classifier in the different regions. The~parameter \(\gamma\) influences the slope of the curve and it is kept constant to 10 to let \(\lambda\) reach the desired value in a suitable number of~epochs.} 
\label{fig:lambda}
\end{figure}

\subsection{Experimental~Settings}
We carried out a complete set of experiments to compare the Transformer encoder classifier performance with and without DANN. {The standard classifier is trained separately on each of the single regions of the dataset, then tested on the other ones. By contrast, DANN models are trained on each source-target pair to gain the desired adaptation capability and tested an all the regions except for the source domain. No validation set is used for model selection. Tests are always performed with the model resulting from a fixed number of training epochs.}

In the final architecture, the~classifier model comprises a transformer encoder feature extractor and a final classification stage. In~all experimentation, the~transformer encoder receives as input a batch of 256 tensors with \(t=45\) temporal steps and \(b=10\) spectral bands in the image samples. Moreover, to~linearly project the temporal sequence to the constant latent dimension of the encoder, the~input is first passed to a dense layer with 64 units. Therefore, $d_{model}$ is equal to 128. On~the other hand, the~multi-head attention Transformer encoder is defined with several layers and attention heads equal to \textit{\(n_{layers} = 3\)} and \textit{\(n_{heads} = 2\)}. Finally, the~dimension of internal fully connected layers \textit{\(d_{inner} = 128\)}. Rectified linear units is the non-linear activation function used for all neurons of the~encoder.

The LC\&CC classification stage is a simple multi-layer perceptron head composed of a normalization layer, a~fully connected layer with 128 units, ReLU as activation function, and~a final layer with $k=9$ neurons.
On the other hand, for~the DANN experimentation, the~domain predictor is identical to the multi-layer perceptron head of the LC\&CC classifier, with~128 units and a ReLU activation. However, the~number of neurons in the final layer is set to \(d = 2\), since we always perform a single target domain~adaptation.

A cross-entropy loss function is chosen to train both the classifiers.
The parameters of both models are updated using Adam optimizer with $\beta_{1}=0.9$, $\beta_{2}=0.999$ and  $\epsilon$ = 1 $\times$ 10$^{-7}$
. A~fixed number of epochs is always set to 250. The~learning rate value is changed during training according to an exponential decay policy from a starting value of 0.001, with~a decay scheduled for each epoch equal to \textit{\(0.99^{epoch}\)}.
A key point in the experimental settings is related to the domain adaptation parameter \(\lambda\). It acts as a regularization parameter, since it regulates the impact of the domain discriminator gradients on the feature extractor during training. Therefore, it can be considered to be the principal hyper-parameter to tune when using DANN. We always use a scheduling policy for \(\lambda\), as~suggested in the original publication of DANN:
\begin{equation}
\label{eqn: lambda_scheduling}
    \lambda_{t} = \lambda_{max} \left(\frac{2}{1+e^{-\gamma t}} -1 \right)
\end{equation}
where \(\lambda_{max}\) is the plateau value reached. This is the actual value of \(\lambda\) used for the second half of the training, which affects the final performance of the model in terms of generalization. The~parameter \(\gamma = 10\) defines the slope of the curve and it is fixed to such value to let \(\lambda_{max}\) be reached in a suitable number of epochs.
A scheduled value of \(\lambda\) allows the feature extractor to learn the basic features for the classification during the first epochs. It then adjusts the mapping function to let the source and target domain feature distributions to overlap at the end of the training process. As~shown in Figure~\ref{fig:lambda}, different values of \(\lambda_{max}\) are tested to study the response of the model. To~our knowledge, \(\lambda_{max} = 0.2\) is the best value for a robust adaptation improvement of the classifier, at~least among the set of tested \(\lambda_{max}\) values.

\begin{figure}[h]
\centering
\includegraphics[scale=0.60]{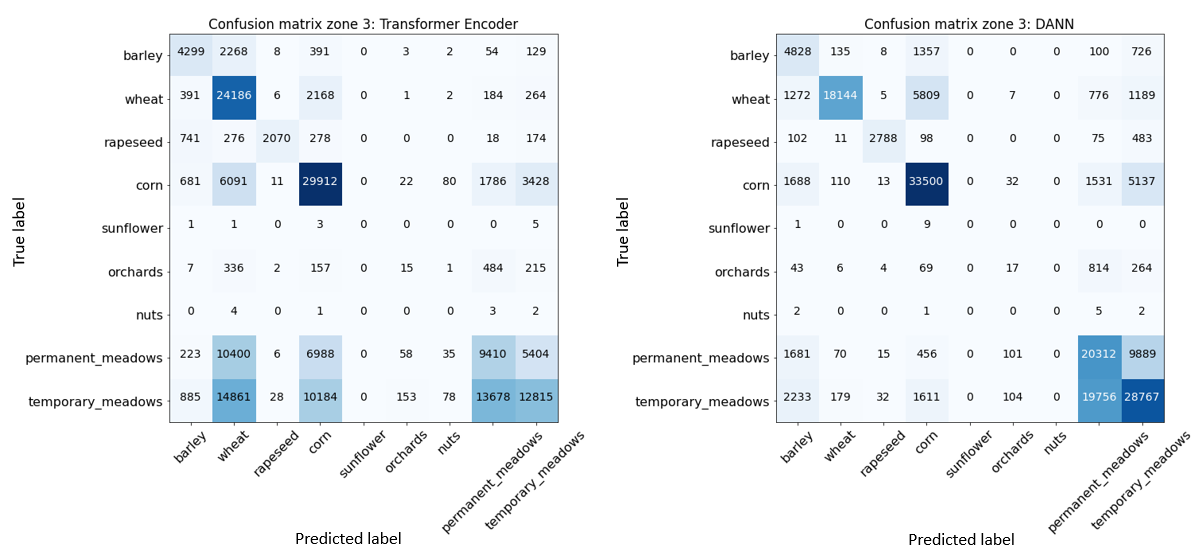}
\caption{{Class-wise comparison of classification results on zone 3 (target), selecting zone 2 as source domain. Confusion matrix obtained with Transformer encoder trained on zone 2 and tested on zone 3 is shown in (\textbf{a}) on the left. Figure (\textbf{b}) on the right shows classification results with DANN model tested on zone 3. The~effect of DANN clearly mitigate the prediction error, with~a particular focus on relevant classes such as Corn, Permanent and Temporary Meadows.}} 
\label{fig:conf_matrix}
\end{figure}

\begin{table}[ht]
\centering
\caption{Results of crops classification for the Transformer Encoder classifier trained with and without DANN using \(\lambda_{max} = 0.2\). The~two models are trained and tested on all the possible combinations of source/target domains available in BreizhCrops dataset. Accuracy, F1-Accuracy and K-score are the metrics used to compare the classification quality. Training accuracy is also reported for the Transformer encoder classifier. Maximum Mean Discrepancy computed on a subset of extracted features of source and target domain shows the successful reduction of features distance obtained with~DANN.}
\label{tab:results1}
\resizebox{\textwidth}{!}{%
\begin{tabular}{ccccccccccc}
\toprule
\multicolumn{2}{c}{\textbf{Zone}} & \multicolumn{5}{c}{\textbf{Transformer Encoder}} & \multicolumn{4}{c}{\textbf{DANN}} \\ \midrule
\begin{tabular}[c]{@{}c@{}}\textbf{Source} \\ \textbf{Domain}\end{tabular} & \begin{tabular}[c]{@{}c@{}}\textbf{Target}\\ \textbf{Domain}\end{tabular} & \begin{tabular}[c]{@{}c@{}}\textbf{Train} \\ \textbf{Accuracy}\end{tabular} & \begin{tabular}[c]{@{}c@{}}\textbf{Test} \\ \textbf{Accuracy}\end{tabular} & \textbf{F1-Accuracy} & \textbf{K-Score} & \textbf{MMD} & \begin{tabular}[c]{@{}c@{}}\textbf{Test }\\ \textbf{Accuracy}\end{tabular} & \textbf{F1-Accuracy} & \textbf{K-Score} &\textbf{ MMD} \\ \midrule
1 & 2 & 0.8577 & 0.7877 & 0.5675 & 0.7229 & 0.1109 & 0.7628 & 0.5540 & 0.6950 & 0.0077 \\
1 & 3 & 0.8577 & 0.7436 & 0.5266 & 0.6606 & 0.1620 & 0.7449 & 0.5080 & 0.6714 & 0.0183 \\
1 & 4 & 0.8577 & 0.7941 & 0.5675 & 0.7294 & 0.0516 & 0.7960 & 0.5734 & 0.7343 & 0.0086 \\
2 & 1 & 0.8951 & 0.7433 & 0.5309 & 0.6773 & 0.1577 & 0.7403 & 0.5161 & 0.6687 & 0.0208 \\
2 & 3 & 0.8951 & 0.4967 & 0.3592 & 0.3642 & 0.6700 & 0.6505 & 0.4544 & 0.5483 & 0.0104 \\
2 & 4 & 0.8951 & 0.6006 & 0.4395 & 0.4912 & 0.2536 & 0.7482 & 0.4832 & 0.6735 & 0.0416 \\
3 & 1 & 0.8750 & 0.7767 & 0.5339 & 0.7122 & 0.1819 & 0.8045 & 0.5778 & 0.7488 & 0.0121 \\
3 & 2 & 0.8750 & 0.6638 & 0.4594 & 0.5615 & 0.6254 & 0.7589 & 0.5334 & 0.6865 & 0.0277 \\
3 & 4 & 0.8750 & 0.7348 & 0.5074 & 0.6504 & 0.1184 & 0.7968 & 0.5778 & 0.7338 & 0.0115 \\
4 & 1 & 0.8870 & 0.7927 & 0.5551 & 0.7354 & 0.0339 & 0.8233 & 0.5822 & 0.7753 & 0.0039 \\
4 & 2 & 0.8870 & 0.7600 & 0.5443 & 0.6870 & 0.0953 & 0.8003 & 0.5788 & 0.7399 & 0.0084 \\
4 & 3 & 0.8870 & 0.7111 & 0.4961 & 0.6230 & 0.0960 & 0.7673 & 0.5443 & 0.6965 & 0.0062 \\ \bottomrule
\end{tabular}%
}
\end{table}

\begin{table}[h]
\centering
\caption{Comparison between Transformer Encoder Classifier with and without DANN, in~terms of classification metrics reported in Table~\ref{tab:results1}. This run of experiments is conducted with a scheduling of the adaptation parameter \(\lambda\), with~\(\lambda_{max} = 0.2\).}
\label{tab:results_improvement1}
\setlength{\tabcolsep}{6.5mm}{
\begin{tabular}{ccrrr}
\toprule
\multicolumn{2}{c}{\textbf{Zone}} & \multicolumn{3}{c}{\textbf{Improvement {[}\%{]}}} \\ \midrule
\begin{tabular}[c]{@{}c@{}}\textbf{Source}\\ \textbf{Domain}\end{tabular} & \begin{tabular}[c]{@{}c@{}}\textbf{Target}\\ \textbf{Domain}\end{tabular} & \multicolumn{1}{c}{\begin{tabular}[c]{@{}c@{}}\textbf{Test}\\ \textbf{Accuracy}\end{tabular}} & \multicolumn{1}{c}{\textbf{F1-Accuracy}} & \multicolumn{1}{c}{\textbf{K-Score}} \\ \midrule
1 & 2 & $-$3.1576 & $-$2.3859 & $-$3.8508 \\
1 & 3 & 0.1762 & $-$3.5378 & 1.6395 \\
1 & 4 & 0.2296 & 1.0467 & 0.6773 \\
2 & 1 & $-$0.3996 & $-$2.7935 & $-$1.2698 \\
2 & 3 & 30.9721 & 26.4916 & 50.5414 \\
2 & 4 & 24.5690 & 9.9474 & 37.1046 \\
3 & 1 & 3.5803 & 8.2152 & 5.1446 \\
3 & 2 & 14.3204 & 16.1075 & 22.2539 \\
3 & 4 & 8.4475 & 13.8791 & 12.8283 \\
4 & 1 & 3.8705 & 4.8817 & 5.4228 \\
4 & 2 & 5.3053 & 6.3384 & 7.6922 \\
4 & 3 & 7.9018 & 9.7154 & 11.8067 \\ \bottomrule
\end{tabular}}
\end{table}

As already explained at the beginning of the section, the~classifiers are trained and tested on all the possible combinations of regions to quantify the existing domain~gap. 

The classification performance is evaluated using three different classification metrics, which are chosen among the ones proposed in the BreizhCrops dataset benchmarks: Accuracy, F1-score and K-score. This last metric is the Cohen’s kappa~\cite{cohen1960coefficient}, computed according \(\kappa = (p_o - p_e) / (1 - p_e)\) where \(p_o\) and \(p_e\) are the empirical and expected probability of agreement on a label.
In addition, we make use of Maximum Mean Discrepancy (MMD) metric, presented in Section~\ref{MMD}, to~quantitatively evaluate the distance between source and target~distributions.

\subsection{Maximum Mean~Discrepancy}
\label{MMD}
MMD is a statistical test originally proposed in~\cite{gretton2012kernel} to determine a measure of the distance between two distributions. MMD is largely used in domain adaptation since it perfectly fits the need to understand whether the source and the target domain extracted features overlap. MMD can be directly exploited as a loss function for adversarial training of generative models or for domain adaptation purposes, as~shown in~\cite{dziugaite2015training,long2017deep}. However, in~this works we limit its usage to show the results of the Transformer Encoder DANN in terms of reduction of feature~distances.\\
Formally, MMD is a kernel-based difference between feature means. Given a set of \(m\) samples \(X\) with a probability measure \(P\), the~feature mean can be expressed as:
\begin{equation}
\label{eqn:feature_mean}
    \mu_{p} ~ (\phi(X)) = \left[ E[\phi(X_{1}], \cdots , E[\phi(X_{m}] \right]^{T}
\end{equation}
where \(\phi(X)\) is the feature map that maps \(X\) to a new feature space \(\mathcal{F}\). If~it satisfies the necessary theoretical conditions, a~kernel-based approach can be used to compute the inner product of two distributions of samples \(X \sim P\) and \(Y \sim Q\):
\begin{equation}
\label{eqn:k_inner_prod}
    \langle \mu_{P}~(\phi(X), \mu_{Q}~(\phi(Y) \rangle_{\mathcal{F}} = E_{P,Q}~ [\langle \phi(X), \phi(Y) \rangle_{\mathcal{F}}] = E_{P,Q}~ [k(X,Y)] 
\end{equation}

At this point the MMD can be defined as the distance between the feature means of \(X \sim P\) and \(Y \sim Q\):
\begin{equation}
\label{eqn:MMD1}
    MMD^{2}(P,Q) = \Vert \mu_{P} - \mu_{Q} \Vert^{2} _{\mathcal{F}}
\end{equation}
which can be expressed more in detail using Equation \eqref{eqn:k_inner_prod}:
\begin{equation}
\label{eqn:MMD2}
    MMD^{2}(P,Q) = E_{P}~ [k(X,X)] - 2 E_{P,Q}~ [k(X,Y)] + E_{Q}~ [k(Y,Y)]
\end{equation}

However, an~empirical estimate of MMD needs to be computed since in a real case only samples are available instead of the explicit formulation of the distributions. It is possible to obtain the MMD expression by considering the empirical estimates of the feature means based on their samples:

\begin{equation}
\label{eqn:MMD_empirical}
    MMD^{2}(X,Y) = \frac{1}{m (m-1)} \sum_{i} \sum_{j \neq i} k(\mathbf{x_{i}}, \mathbf{x_{j}}) - 2 \frac{1}{m.m} \sum_{i} \sum_{j} k(\mathbf{x_{i}}, \mathbf{y_{j}}) + \frac{1}{m (m-1)} \sum_{i} \sum_{j \neq i} k(\mathbf{y_{i}}, \mathbf{y_{j}})
\end{equation}

\noindent where \(\mathbf{x_{i}}\) and \(\mathbf{y_{i}}\) in this case are the image samples from source and target domains, \(m\) is the number of samples of the considered subsets.
Finally, we specifically use a gaussian kernel with the following expression:
\begin{equation}
\label{eqn:radial_kernel}
    k(\mathbf{x_{i}}, \mathbf{x_{j}}) = \exp \left(\frac{- \Vert \mathbf{x_{i}} - \mathbf{x_{j}} \Vert^{2}}{2\sigma^{2}}\right) = \exp \left(\frac{-1}{\sigma^{2}} [\mathbf{x_{i}}^\intercal \mathbf{x_{i}} - 2 \mathbf{x_{i}}^\intercal \mathbf{x_{j}} + \mathbf{x_{j}}^\intercal \mathbf{x_{j}}]\right)
\end{equation}

\begin{figure}[ht]
\includegraphics[scale=0.75]{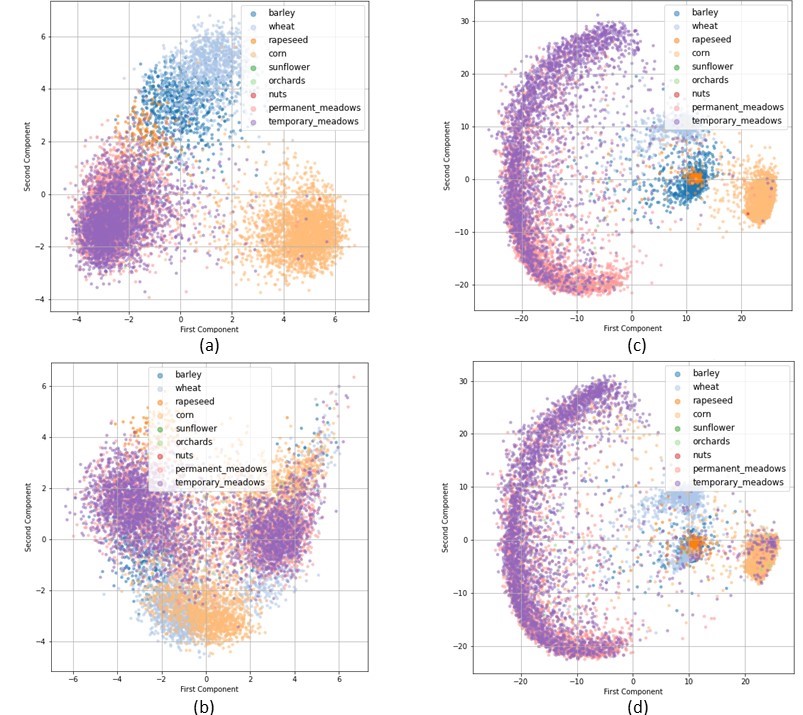}
\caption{{2D feature}~visualization obtained with PCA, extracted with the Transformer Encoder trained on the source domain and with the Transformer DANN model trained on the specific source-target domains. {A comparison between the 2D feature distributions is proposed for the case of zone 2 (source) and 3 (target). In~(\textbf{a},\textbf{b}) we have features extracted with the Transformer Encoder from source and target domains: (\textbf{a}) reports features of the source domain (zone 2) and (\textbf{b}) the ones extracted from the target domain (zone 3). In~this case, features are mapped poorly in the target domain, with~a consequent low accuracy in classification. In~(\textbf{c},\textbf{d}) the same features extracted with the Transformer DANN model are shown. The~positive effect of DANN in terms of features overlapping is evident compared to (\textbf{a},\textbf{b}).}} 
\label{fig:2D_PCA_comparison_2_3}
\end{figure}

\begin{figure}[ht]
\includegraphics[scale=0.75]{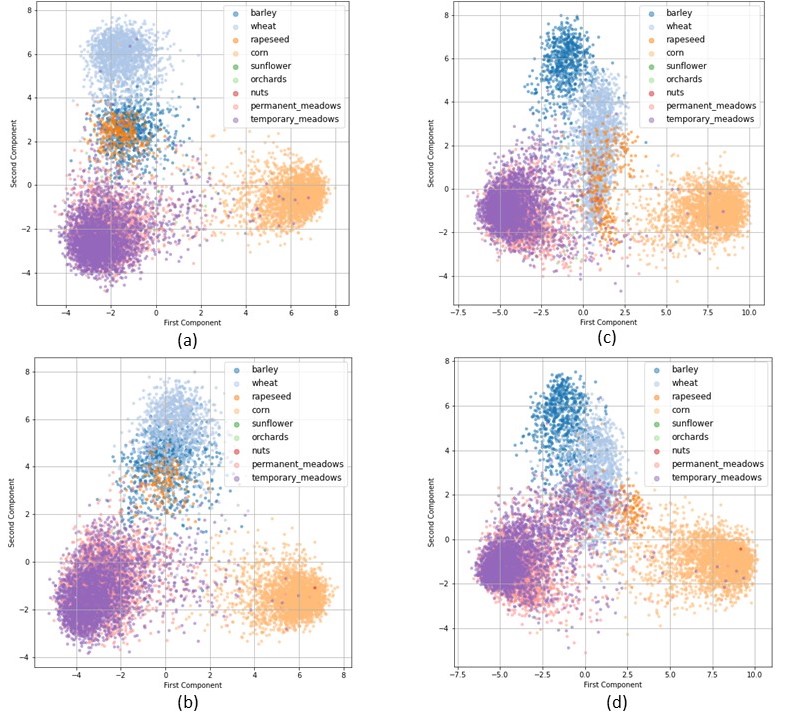}
\caption{{2D feature}~visualization obtained with PCA, extracted with the Transformer Encoder trained on the source domain and with the Transformer DANN model trained on the specific source-target domains. {A comparison between the 2D feature distributions is proposed for the case of zone 1 (source) and 2 (target). In~(\textbf{a},\textbf{b}) we have features extracted with the Transformer Encoder from source, (\textbf{a}), and~target, (\textbf{b}), domain: a low MMD distance indicates no need for domain adaptation. In~(\textbf{c},\textbf{d})~the same features extracted with the Transformer DANN model are shown, with~no substantial differences.}} 
\label{fig:2D_PCA_comparison_1_2}
\end{figure}

\begin{figure}[ht]
\includegraphics[scale=0.75]{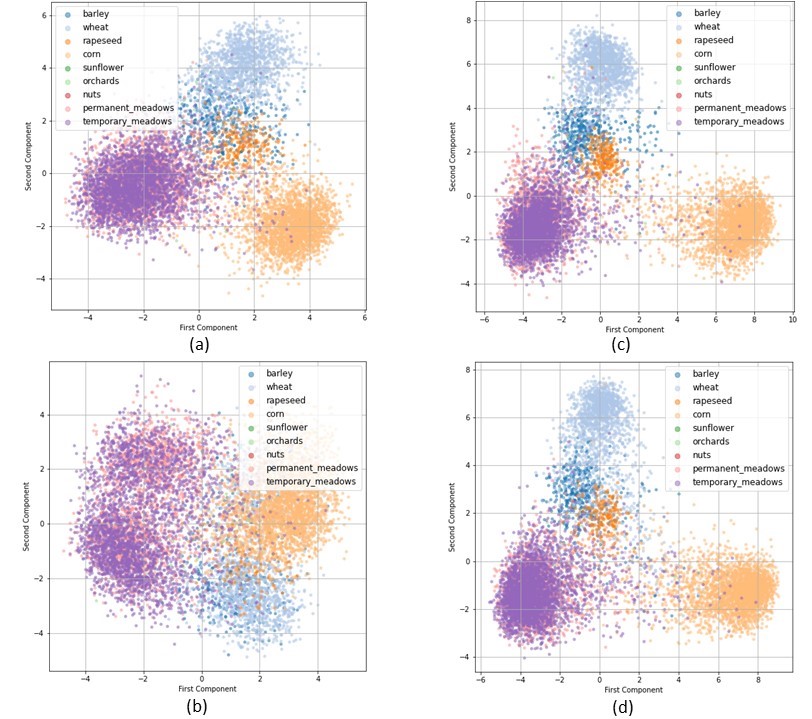}
\caption{2D feature visualization obtained with PCA, extracted with the Transformer Encoder trained on the source domain and with the Transformer DANN model trained on the specific source-target domains. {A comparison between the 2D feature distributions is proposed for the case of zone 4 (source) and 3 (target). In~(\textbf{a},\textbf{b}) we have features extracted with the Transformer Encoder from source, (\textbf{a}), and~target, (\textbf{b}), domains: regardless of an initial low MMD, the~classifier accuracy can still be improved reducing the domain gap. In~(\textbf{c},\textbf{d}) the same features extracted with the Transformer DANN model are shown, with~a clear improvement of the feature mapping, which result in very similar distributions from source to target~domain.}}
\label{fig:2D_PCA_comparison_4_3}
\end{figure}

\subsection{Results Discussion and Applicability~Study}
In this section, we present the comparison results between the Transformer classifier with and without DANN, clearly highlighting the scenarios that present a definite advantage in applying adversarial training for training a classifier for LC\&CC.
From results in Tables~\ref{tab:results1} and \ref{tab:results_improvement1}, Figure~\ref{fig:conf_matrix}, it is possible to notice that DANN adversarial training allows the classifier to improve knowledge transferability to other domains for most of the cases. Nonetheless, we investigate a potential criterion to decide if the transfer of learning from source to target can be effectively improved by DANN. More in detail, since DANN aims to overlap feature distributions, we look at the extracted features from a subset of 10000 samples of each zone dataset that is considered representative of the total~one.

We use the set of extracted features to compute a numerical evaluation of the distance decrease, and~to give a graphical visualization of the effect of DANN. From~a quantitative perspective, we propose Maximum Mean Discrepancy as the feature distance metrics to detect suitable conditions where DANN is an appropriate methodology. To~compute MMD without considering the clustering of classes, we only need unlabeled image samples. We use PCA algorithm to compute the principal components of the extracted features and we exploit them to provide 2D and 3D visualization of relevant~cases.

First, we can look at the MMD values obtained from both the Transformer encoder and DANN in Table~\ref{tab:results1}. It is clear that DANN is always able to reduce the distance between feature distributions. However, this is not always associated with an increase in classification performance. We realize that key information is contained in the MMD value obtained from source and target features, extracted by the standard classifier. This simple test is crucial and can also be done without labels. The~best improvement with DANN is reached considering zone 2 as the source domain and selecting zone 3 as the target domain. The~percentage improvement shown in Table~\ref{tab:results_improvement1}, with~an increase of more than \(30\%\) of accuracy, correlates with an initial MMD value for this specific case is equal to 0.6700, reduced by DANN to 0.0104. What can be deduced by this observation is  that high values of the MMD indicate a lack of generalization of the classifier and a domain gap. It is also to consider that the geographical zones of interest are close to each other. Hence, it can be reasonable to find small domain gaps. A~clear example is the case of zone 1, when chosen as source domain. This factor can be considered an additional difficulty of the study case. Therefore, it is possible that the same methodology applied to other regions on the planet, sharing the same categories of crops, can probably show greater results. Another peculiar case to be considered is: zones 4 (source) and 3 (target). The MMD value is low from the initial analysis of the case, without~the intervention of DANN. However, a classification boost is always achieved.

We report a visual representation of the extracted features to add meaning to the previous considerations. In~particular, Figures~\ref{fig:2D_PCA_comparison_2_3}--\ref{fig:2D_PCA_comparison_4_3} show the 2D principal components obtained from the peculiar cases defined~below:
\begin{itemize}
    \item \textbf{case 1}: zone 2 (source), zone 3 (target). In~this case DANN shows the greatest improvements with an initial high value of MMD. Features are visually reported in \mbox{Figure~\ref{fig:2D_PCA_comparison_2_3}}: in (a,b) when extracted by standard Transformer encoder trained on the source domain, in~(c,d) when extracted by DANN. The~difference is visually clear. Features distributions are matched by DANN, with~a resulting overlapping shape between source and target domain.
    \item \textbf{case 2}: zone 1 (source), zone 2 (target). In~this case DANN shows the worst improvements with an initial low value of MMD. Features are visually reported in (a,b) of Figure~\ref{fig:2D_PCA_comparison_1_2} when extracted by standard Transformer encoder, in~(c,d) of the same \mbox{Figure~\ref{fig:2D_PCA_comparison_1_2}} when extracted by DANN. They appear already similar also without DANN.
    \item \textbf{case 3}: zone 4 (source), zone 3 (target). In~this case DANN shows noticeable improvements, regardless an initial low value of MMD. Features are visually reported in (a,b) of Figure~\ref{fig:2D_PCA_comparison_4_3} when extracted by standard Transformer encoder, in~(c,d) of Figure~\ref{fig:2D_PCA_comparison_4_3} when extracted by DANN. As with case 1, the~difference is visually clear, and the effect of DANN can be easily appreciated.
\end{itemize}

\begin{figure}[ht]
\centering
\includegraphics[scale=0.77]{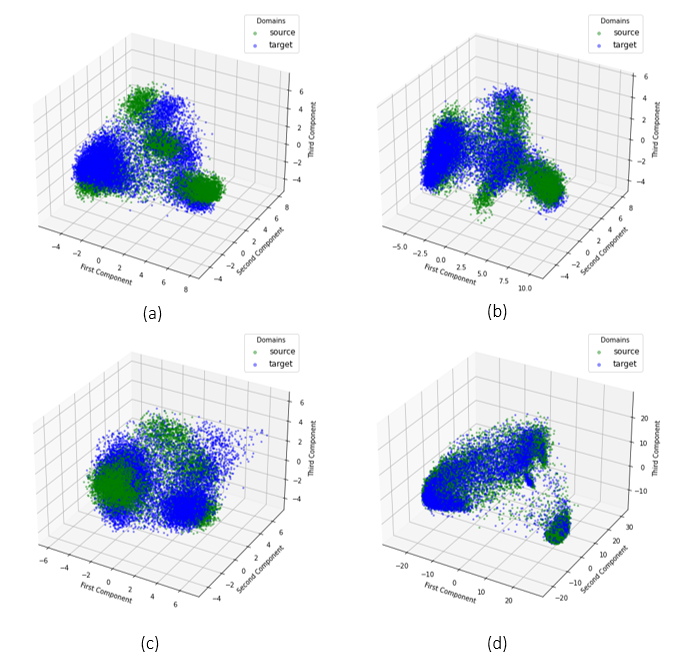}
\caption{3D feature visualization and comparison. (\textbf{a},\textbf{b}) show the features extracted from zone 1 (source) and 2 (target). They are respectively obtained with transformer encoder and DANN. It is clear that the transformer encoder alone can correctly map features on both domains. By contrast, the~improvement provided by DANN model is very evident in figures (\textbf{c},\textbf{d}), representing the features extracted from zone 2 (source) and 3 (target), where the transformer encoder alone present both high values of MMD and low classification accuracy on target~domain.} 
\label{fig:3D_PCA}
\end{figure}

Finally, case 1 and case 2 defined above are also considered for a 3D representation. Figure~\ref{fig:3D_PCA} shows the obtained results. For~each subplot in the figure, both source and target domain features are scattered. Thanks to this visual perspective, the~effect of the DANN method is highlighted, considering both the worst and the best application scenario. In~case 1, the~difference between source and target features is shallow also without DANN, as~shown in (a). By contrast, the~situation from (c) to (d) is changed thanks to the adversarial training~significantly.

The proposed discussion underlines some interesting insights on the correlation between reducing the domain gap and improving a classifier performance. The~isolated cases considered provide a good reference example to decide if it is a reasonable and convenient choice to adopt the proposed DANN methodology for multi-spectral temporal sequences for Land Cover~classification. 

\section{Conclusions}
\label{conclusion}
In this paper, we investigated adversarial training for domain adaptation with state-of-the-art self-attention-based models for LC\&CC. Indeed, domain gaps between distinct geographical regions prevent the direct repurpose of the trained model on diverse areas of the training domain, and~the practical difficulty of acquiring labeled data prevents the direct application of transfer learning techniques. Our extensive experimentation clearly highlights the advantages of applying the proposed methodology to transformer models trained on multi-spectral, multi-temporal data and the considerable gain in performance with considerable distribution distance between target and source regions. {In particular, the~best improvement obtained with DANN shows a percentage increase of more than} \(30\%\) {of classification accuracy, associated with an evident reduction of the features distance metrics MMD from 0.6700 to 0.0104. Moreover, our investigations conduct to a clear identification of the scenarios where it is advantageous to apply the DANN domain adaptation mechanism. More in detail we identified three different cases that highlight the strategy for a correct adoption of the methodology. A~graphical visualization of the effect of DANN on the crop classification task has also been proposed and discussed exploiting the 2D class-wise and the 3D principal components of crops features distribution.}

Future work may investigate the advantages and disadvantages of different domain adaptation techniques applied to LC\&CC {and extend our study to further geographical~regions}.

\bibliographystyle{unsrt}  
\bibliography{bibliography}  







\end{document}